%% file: main.tex
\documentclass{article}

\usepackage{microtype}
\usepackage{graphicx}
\usepackage{subcaption}
\usepackage{booktabs} 

\usepackage{hyperref}

\usepackage[accepted]{icml2026}

\makeatletter
\renewcommand{\printAffiliationsAndNotice}[1]{%
  \global\icml@noticeprintedtrue%
  {\let\thefootnote\relax\footnotetext{%
    \hspace*{-\footnotesep}%
    \textsuperscript{1}Hunan University. %
    Correspondence to: Xiao Wang \textless{}wang0l1@hnu.edu.cn\textgreater{}, %
    Changjian Chen \textless{}changjianchen@hnu.edu.cn\textgreater{}, %
    Zhuo Tang \textless{}ztang@hnu.edu.cn\textgreater{}.
  }}%
}
\makeatother

\usepackage{amsmath}
\usepackage{amssymb}
\usepackage{mathtools}
\usepackage{amsthm}
\usepackage[capitalize,noabbrev]{cleveref}

\theoremstyle{plain}

\theoremstyle{definition}

\theoremstyle{remark}

\usepackage[textsize=tiny]{todonotes}
\icmltitlerunning{StationPDE: Station-Oriented Surface PDE Learning for Multi-Station Multivariate Weather Forecasting}
\usepackage{enumitem}

\usepackage{amsfonts}
\usepackage{multirow}
\usepackage{makecell}
\usepackage{pifont}

\newcommand{\method}{StationPDE}

\def \eg {{\emph{e.g}.}}

\newcommand{\myparagraph}[1]{\par\noindent\textbf{#1}\hspace{0.5em}\ignorespaces}
\newcommand{\best}[1]{\textbf{#1}}
\newcommand{\second}[1]{\underline{#1}}
\usepackage{type1cm}
\newcommand{\std}[1]{\scalebox{0.68}{$_{\pm #1}$}}

\begin{document}

\twocolumn[
  \icmltitle{StationPDE: Station-Oriented Surface PDE Learning for Multi-Station Multivariate Weather Forecasting}
  \icmlsetsymbol{equal}{*}
    \begin{icmlauthorlist}
    \icmlauthor{Xiao Wang}{hnu}
    \icmlauthor{Changjian Chen}{hnu}
    \icmlauthor{Rongwen Li}{hnu}
    \icmlauthor{Hongwu Liu}{}
    \icmlauthor{Kun Fang}{}
    \icmlauthor{Zhuo Tang}{hnu}
    \end{icmlauthorlist}

    \icmlaffiliation{hnu}{Hunan University}

    \icmlcorrespondingauthor{Xiao Wang}{wang0l1@hnu.edu.cn}
    \icmlcorrespondingauthor{Changjian Chen}{changjianchen@hnu.edu.cn}
    \icmlcorrespondingauthor{Zhuo Tang}{ztang@hnu.edu.cn}
]

\printAffiliationsAndNotice{}  

\input{paper/0_abstract}  
\input{paper/1_Introduction} 
\input{paper/2_Related_Work} 
\input{paper/3_Problem_Formulation} 
\input{paper/4_Framework} 
\input{paper/5_Experiments} 
\input{paper/6_Conclusion}

\bibliography{main}
\bibliographystyle{icml2026}

\end{document}

%% file: paper/0_abstract.tex
\begin{abstract}
Multi-station multivariate weather forecasting aims to forecast future weather variables at multiple weather stations from historical surface observations.
Existing station forecasting models learn statistical dependencies among discrete stations, but lack explicit physical evolution.
Meanwhile, PDE-based weather models provide interpretable physical dynamics, yet require continuous fields and upper-air variables unavailable in surface station data.
To bridge this gap, we propose \method{}, a station-oriented surface PDE learning model.
\method{} constructs a terrain-aware continuous surface field from discrete station observations and decomposes its physical evolution into surface wind transport and upper-air inference.
Surface wind transport explicitly evolves observable weather variables, while upper-air inference uses learnable horizontal diffusion to approximate the missing influence of unavailable upper-air variables.
A parallel data-driven diffusion branch captures complementary motion patterns, and an adaptive router integrates the two forecasts for station-level multivariate forecasting.
Experiments on Weather2K and MeteoNet show that \method{} consistently outperforms state-of-the-art baselines, reducing MSE by about $9.6\%$ on average compared with the strongest baseline.
\end{abstract}

%% file: paper/1_Introduction.tex
\section{Introduction}
Multi-station multivariate weather forecasting forecasts future weather variables at multiple weather stations from their historical observations.
It is widely used in weather-sensitive applications, such as local weather services, risk management, and renewable energy scheduling~\cite{zhang2023skilful}.
Existing multi-station weather forecasting models typically adopt data-driven strategies to model inter-station dependencies~\cite{Cao2020StemGNN, han2023MasterGNN, wu2023Corrformer}. 
While these models can capture statistical correlations among observed stations, they usually characterize weather evolution through discrete station-to-station relationships, which may fail to reflect the continuous evolution of the underlying weather process.
A recent model, CDPNet~\cite{xu2025CDPNet}, moves beyond discrete station modeling by constructing a spatially continuous field from station observations and estimating diffusive differences over time.
However, their evolution process remains largely data-driven and lacks explicit modeling of multivariate physical coupling.

\begin{figure}[t]
    \centering
    \includegraphics[width=\linewidth]{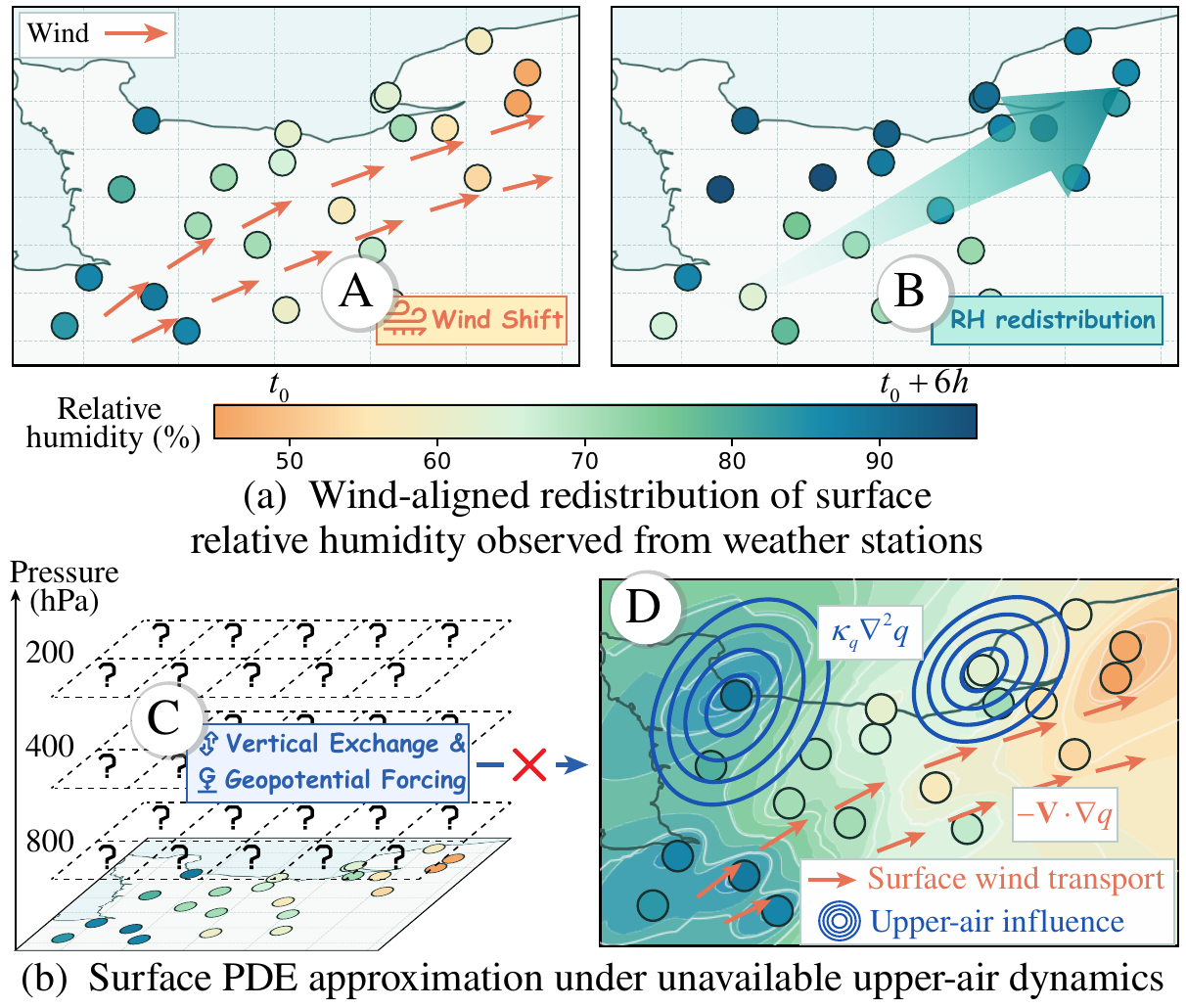}
    \caption{
    Motivation of station-oriented surface PDE modeling.
    (a) Wind-aligned redistribution of surface relative humidity observed from MeteoNet stations.
    (b) Surface PDE approximation under unavailable upper-air dynamics, where surface wind transport and upper-air influence are modeled on the station-derived surface field.
    }
    \label{fig:intro_motivation}
\end{figure}

Figure~\ref{fig:intro_motivation}(a) illustrates why such multivariate physical coupling is important in multi-station multivariate weather forecasting.
In a real case from the MeteoNet dataset~\cite{meteonet2020} in northwestern France, future surface relative humidity is not solely determined by each station's own history.
Instead, the dominant surface wind at $t_0$ (Fig.~\ref{fig:intro_motivation}A) is followed by a clear redistribution of relative humidity after six hours (Fig.~\ref{fig:intro_motivation}B), suggesting a wind-driven transport process across stations.
This observation indicates that weather forecasting requires more than learning which stations are statistically related; it also requires modeling how weather variables are transported over space.

A natural way to introduce such physical evolution is to embed atmospheric partial differential equation (PDE) processes into weather forecasting models.
For example, WeatherGFT~\cite{xu2024weathergft} incorporates PDE-based processes into the forward pass and demonstrates the value of explicit physical dynamics for weather forecasting.
However, existing PDE-based weather models are mainly designed to evolve continuous states on regular continuous fields, whereas surface station observations and available only at discrete locations~\cite{Zhu2023weather2K,meteonet2020}.
Moreover, standard atmospheric PDE formulations often rely on multi-pressure-level upper-air variables, such as geopotential and vertical velocity, while surface station datasets only have surface observations (\eg, Fig.~\ref{fig:intro_motivation}C).
This leaves a clear gap: \textbf{multi-station weather forecasting models lack explicit physical evolution, while existing PDE-based weather models cannot be directly applied when only surface station observations are available.}

To bridge this gap, we propose \method{}, a station-oriented surface PDE learning model for multi-station multivariate weather forecasting.
\method{} first constructs a continuous surface field from discrete station observations through terrain-aware surface field initialization.
To address the incomplete specification of physical evolution caused by missing upper-air variables, \method{} decomposes the evolution process into two complementary parts: \emph{surface wind transport} and \emph{upper-air inference}.
For \emph{surface wind transport}, since the related surface variables are available in station observations, it directly incorporates these variables, including thermodynamic state, pressure transport, humidity evolution, and wind dynamics, in the PDE.
For \emph{upper-air inference}, since the related variables are not available, we use learnable horizontal diffusion to approximate their missing influence on the surface field, enabling local weather changes to propagate across neighboring regions (\eg, Fig.~\ref{fig:intro_motivation}D).
Together, surface wind transport and horizontal diffusion form a complete PDE system for modeling the physical evolution of the surface field.
A parallel data-driven diffusion branch further captures residual motion patterns, and an adaptive router integrates the two forecasts and maps the fused continuous field back to station-level multivariate forecasts.

In summary, our contributions are as follows.
\begin{itemize}
    \item \textbf{A station-oriented surface PDE learning model}, which introduces explicit physical evolution into multi-station multivariate weather forecasting using only surface station observations.
    
    \item \textbf{A surface PDE evolution design}, which decomposes surface field evolution into surface wind transport and upper-air inference, making explicit physical evolution feasible from surface station observations.
    
    \item \textbf{Extensive experiments on real-world datasets}, which show that our model outperforms state-of-the-art baselines on Weather2K and MeteoNet, reducing MSE by about $9.6\%$ on average compared with the strongest baseline.
\end{itemize}

%% file: paper/2_Related_Work.tex
\section{Related Work}
\label{sec:related}
\subsection{Data-Driven Multi-Station Forecasting Models}
Data-driven multi-station forecasting models mainly learn temporal patterns and inter-station dependencies from historical observations.
For example, Autoformer~\cite{wu2021autoformer} replaces standard self-attention with an Auto-Correlation mechanism and seasonal-trend decomposition for long-term forecasting.
Building on Autoformer, Corrformer~\cite{wu2023Corrformer} introduces a multi-correlation mechanism that unifies spatial cross-correlation and temporal auto-correlation for regional or global forecasting.
Formulating multi-station weather forecasting as a graph learning problem, MasterGNN+~\cite{han2023MasterGNN} learns multi-view station graphs from geographic distance and environmental context.

Recent general multivariate time series models complement multi-station models by improving variable-wise dependency modeling.
iTransformer~\cite{liu2024itransformer} applies Transformer components on inverted dimensions, embedding each variate's historical series as a token to capture multivariate correlations.
TimeXer~\cite{wang2024timexer} models forecasting with exogenous variables, reconciling endogenous and exogenous information through patch-wise self-attention and variate-wise cross-attention.

However, these models mainly operate on observed sequences rather than constructing continuous weather fields from discrete stations.
Recently, CDPNet~\cite{xu2025CDPNet} moves beyond purely discrete station modeling by lifting discrete station observations into continuous fields.
Nevertheless, its evolution process remains largely data-driven and does not explicitly characterize the physical coupling among weather variables.
In contrast, \method{} constructs a station-derived surface continuous field and evolves it with station-oriented surface PDE processes.
\begin{figure*}[t]
    \centering
    \includegraphics[width=\linewidth]{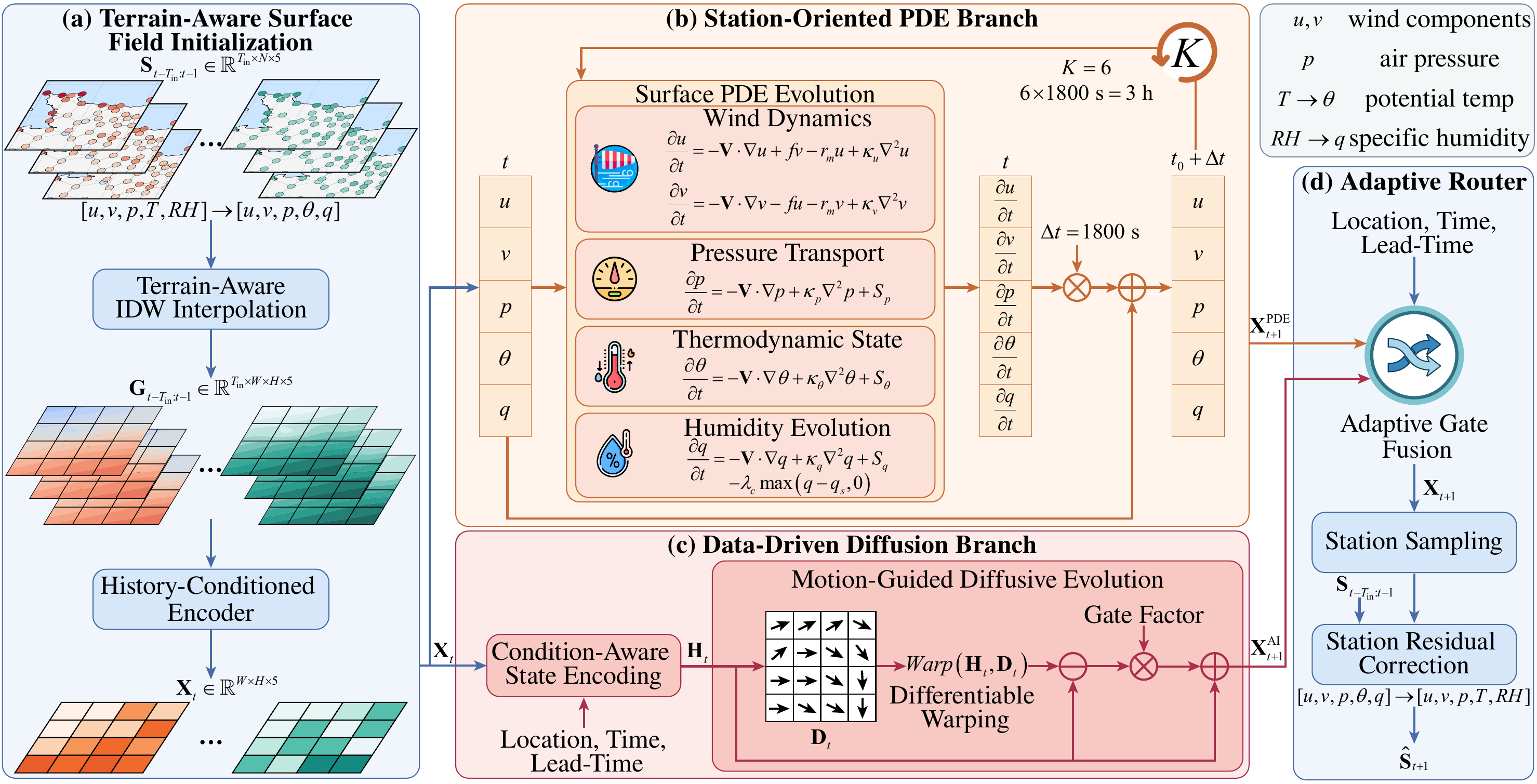}
    \caption{The overview of the \method{} structure.}
    \label{fig:system}
\end{figure*}
\subsection{Physical-Evolution Weather Forecasting Models}
Physical knowledge has been increasingly introduced into weather forecasting models to improve interpretability and physical consistency.
For example, DGFormer~\cite{xu2024dgformer} is a physics-guided station-level weather forecasting model that uses a continuously varying graph topology and inserts domain knowledge into spatial-temporal graph learning.
For discrete station observations, physics-guided weather reconstruction ~\cite{soto2024physics} reconstructs dense wind and pressure fields from discrete station measurements under Navier-Stokes regularization.
PhyDL-NWP~\cite{luo2025PhyDL} computes physical terms by automatic differentiation and uses physics-informed losses with latent force parameterization to guide continuous weather modeling.
However, these models mainly use physical knowledge as graph guidance, loss regularization, or parameterized correction, rather than explicit station-oriented PDE forward evolution over weather variables.

Recently, WeatherGFT~\cite{xu2024weathergft} embeds physical evolution directly into the model forward process, using a PDE kernel for fine-grained physical evolution and a parallel neural branch for adaptive correction.
However, such PDE-based weather models are usually designed for regular continuous fields and rely on upper-air variables unavailable in surface station observations.
In contrast, \method{} performs surface PDE evolution over observable weather variables on a station-derived continuous field, enabling PDE-guided multi-station forecasting from surface station data.

%% file: paper/3_Problem_Formulation.tex
\section{Problem Formulation}
\label{sec:problem}
We study multi-station multivariate weather forecasting from surface station observations.
Consider $N$ weather stations distributed in a region.
Each station has fixed geographic information, including longitude, latitude, and altitude.
We denote the station geographic information as $\mathbf{L}^{\mathrm{station}}\in\mathbb{R}^{N\times 3}$.

At each time step $t$, the observable weather variables at all stations are denoted by
$\mathbf{S}_t\in\mathbb{R}^{N\times C}$.
In this work, we use $C=5$ observable variables ordered as
$\mathbf{S}_t=[u,v,p,T,RH]$, where $p$, $T$, and $RH$ denote air pressure, air temperature, and relative humidity, respectively, and wind speed and direction are converted into horizontal components $u$ and $v$ along the eastward and northward directions.
Given the historical station observations over the past $T_{\mathrm{in}}$ time steps,
$\mathbf{S}_{t-T_{\mathrm{in}}:t-1}\in\mathbb{R}^{T_{\mathrm{in}}\times N\times C}$,
the goal is to generate future station-level weather forecasts for the next $T_{\mathrm{out}}$ time steps:
\begin{equation}
    \mathbf{S}_{t-T_{\mathrm{in}}:t-1}, \mathbf{L}
    \longrightarrow
    \hat{\mathbf{S}}_{t:t+T_{\mathrm{out}}-1},
\end{equation}
where
$\hat{\mathbf{S}}_{t:t+T_{\mathrm{out}}-1}
\in\mathbb{R}^{T_{\mathrm{out}}\times N\times C}$.

In this work, we focus on introducing explicit physical evolution into this station-based forecasting setting.
\method{} lifts the station observations $\mathbf{S}$ into a station-derived surface continuous field, evolves the field with station-oriented surface PDE processes, and maps the evolved field back to station-level forecasts.
To represent this continuous field, we discretize the study region into a regular $H\times W$ grid, where $H$ and $W$ denote the numbers of grid cells along the latitude and longitude directions, respectively, and $M=HW$.
We denote the geographic information of all grid cells by
$\mathbf{L}^{\mathrm{grid}}\in\mathbb{R}^{M\times 3}$.

%% file: paper/4_Framework.tex
\section{The \method{} Framework}
\label{sec:methods}
In this section, we introduce the proposed \method{} model for multi-station multivariate weather forecasting from surface station observations.
As shown in Fig.~\ref{fig:system}, \method{} consists of four key components:
(1) a \textbf{Terrain-Aware Surface Field Initialization} module that lifts discrete station observations into a station-derived surface continuous field;
(2) a \textbf{Station-Oriented PDE Branch} that models surface field evolution by combining surface wind transport with upper-air inference over observable weather variables;
(3) a \textbf{Data-Driven Diffusion Branch} that captures complementary motion and correction patterns beyond the simplified surface PDEs;
and (4) an \textbf{Adaptive Router} that integrates the PDE-based and data-driven forecasts and maps the fused field back to station-level multivariate forecasts.
The following subsections describe these components in detail.

\subsection{Terrain-Aware Surface Field Initialization}
\label{sec:terrain_init}
The first step of \method{} is to construct a surface continuous field from discrete and irregular station observations.
Before spatial lifting, we convert observable variables
$\mathbf{S}_t=[u,v,p,T,RH]$ into physical state channels
$\Psi(\mathbf{S}_t)=[u,v,p,\theta,q]$, where $\theta$ and $q$ are potential temperature and specific humidity.
The inverse transformation $\Psi^{-1}(\cdot)$ maps the physical state back to observable variables, with details provided in the appendix.

CDPNet~\cite{xu2025CDPNet} shows that interpolating station observations over latitude and longitude can construct a continuous spatial field from discrete stations.
However, such location-only interpolation ignores terrain effects, which is problematic for surface variables such as temperature and pressure.
Therefore, \method{} performs terrain-aware surface field initialization with inverse distance weighting (IDW) interpolation, in which both horizontal location and altitude are used to compute interpolation weights.
Let $\mathbf{L}^{\mathrm{station}}_{i,:}$ and
$\mathbf{L}^{\mathrm{grid}}_{m,:}$ denote the geographic information of station $i$ and grid cell $m$, respectively.
For $c\in\{\mathrm{lon},\mathrm{lat},\mathrm{alt}\}$, we define the coordinate difference as
$\Delta l^{c}_{mi}=L^{\mathrm{grid}}_{m,c}-L^{\mathrm{station}}_{i,c}$.
The grid altitude is first estimated from station altitudes by horizontal IDW.
For each grid cell $m$, we select a neighboring station set $\mathcal{N}(m)$ and compute the terrain-aware distance between station $i$ and grid cell $m$ as
\begin{equation}
\begin{aligned}
d_{mi}
=
\Big[
&\big(R\cos(\bar{l}_{\mathrm{lat}})
\Delta l^{\mathrm{lon}}_{mi}\big)^2
+
\big(R\Delta l^{\mathrm{lat}}_{mi}\big)^2 \\
&+
\big(w_{\mathrm{alt}}\Delta l^{\mathrm{alt}}_{mi}\big)^2
\Big]^{1/2},
\quad i\in\mathcal{N}(m),
\end{aligned}
\end{equation}
where $R$ is the Earth radius, $\bar{l}_{\mathrm{lat}}$ is the mean latitude of the study region, and $w_{\mathrm{alt}}=5$ controls the contribution of altitude to the interpolation distance.

Based on this distance, the terrain-aware IDW coefficient from station $i$ to grid cell $m$ is defined as
\begin{equation}
    \alpha_{mi}
    =
    \frac{(d_{mi}+\epsilon)^{-1}}
    {\sum_{j\in\mathcal{N}(m)}(d_{mj}+\epsilon)^{-1}},
    \quad i\in\mathcal{N}(m),
\end{equation}
where $j$ indexes neighboring stations for normalization and $\epsilon$ is a small constant for numerical stability.
Each historical station state is then lifted to grid cell $m$ by
\begin{equation}
    \mathbf{G}_{\tau,m}
    =
    \sum_{i\in\mathcal{N}(m)}
    \alpha_{mi}
    [\Psi(\mathbf{S}_{\tau})]_i,
    \quad
    \tau=t-T_{\mathrm{in}},\ldots,t-1 .
\end{equation}
This produces a historical surface continuous field
$\mathbf{G}_{t-T_{\mathrm{in}}:t-1}
\in\mathbb{R}^{T_{\mathrm{in}}\times C\times H\times W}$.

PDE evolution is driven by an initial spatial state.
If we directly use the latest interpolated field $\mathbf{G}_{t-1}$, the temporal information in the historical window may be underused.
We therefore introduce a history-conditioned encoder.
The historical tensor is concatenated with initialization condition fields $\mathbf{U}_t$, including location and historical time features, to generate a candidate initial field:
\begin{equation}
    \widetilde{\mathbf{X}}_{t}
    =
    \mathcal{E}_{\phi}
    \big(
    [\mathbf{G}_{t-T_{\mathrm{in}}:t-1},\mathbf{U}_{t}]
    \big),
\end{equation}
where $\mathcal{E}_{\phi}$ denotes the history-conditioned encoder, implemented with lightweight convolutional layers to aggregate the historical sequence and condition fields.
The initial field is then obtained by a gated correction:
\begin{equation}
    \mathbf{X}_{t}
    =
    \mathbf{G}_{t-1}
    +
    \mathbf{M}_{t}^{\mathrm{init}}\odot
    (\widetilde{\mathbf{X}}_{t}-\mathbf{G}_{t-1}),
\end{equation}
where $\mathbf{M}_{t}^{\mathrm{init}}$ is a sigmoid initialization gate generated from the latest continuous field, encoded history features, and initialization condition fields.
The resulting $\mathbf{X}_{t}\in\mathbb{R}^{C\times H\times W}$ serves as the initial surface field for both the station-oriented PDE branch and the data-driven diffusion branch.

\subsection{Station-Oriented PDE Branch}
\label{sec:pde_branch}
The station-derived surface continuous field $\mathbf{X}_t$ provides a spatial state for PDE evolution, but it only contains surface observable variables rather than full atmospheric states.
Thus, standard atmospheric PDE formulations that rely on upper-air variables, such as vertical velocity and geopotential, cannot be directly applied.
To address the incomplete specification of physical evolution caused by missing upper-air variables, \method{} decomposes surface field evolution into two complementary parts: \emph{surface wind transport} and \emph{upper-air inference}.

\emph{Surface wind transport}.
Let $\mathbf{V}=(u,v)$ denote the surface horizontal wind field, and let $\nabla=(\partial_x,\partial_y)$ denote the horizontal gradient operator.
Following standard atmospheric PDE formulations, \emph{surface wind transport} is modeled by the surface computable term $-\mathbf{V}\cdot\nabla a$ for each state variable $a\in\{u,v,p,\theta,q\}$.
This term describes how observable weather variables are transported over the continuous surface field by horizontal wind.

\emph{Upper-air inference}.
We consider the missing influence of vertical momentum transport and geopotential gradient forcing, both of which require upper-air variables unavailable in surface station observations.
Without these terms, the surface field lacks part of the large-scale spatial forcing needed to propagate local weather changes across neighboring regions.
We therefore introduce learnable horizontal diffusion for each state variable.
Its Laplacian form spreads local anomalies over the surface field, providing a surface-level approximation to the missing upper-air influence.
This design is inspired by the explicit spatial diffusion treatment in the WRF-ARW technical note~\cite{skamarock2019description}, where second-order horizontal diffusion is introduced for model variables, with corresponding forms for both momentum and scalar variables.

Accordingly, for each state variable $a$, we use the following transport-diffusion form:
\begin{equation}
    \frac{\partial a}{\partial t}
    =
    -\mathbf{V}\cdot\nabla a
    +
    \kappa_a\nabla^2 a
    +
    \mathcal{R}_a ,
\end{equation}
where $\kappa_a$ is the horizontal diffusion coefficient, and $\mathcal{R}_a$ collects variable-specific surface forcing or closure terms.

For \emph{wind dynamics}, $\mathcal{R}_a$ includes the Coriolis force and surface friction:
\begin{equation}
\begin{aligned}
\frac{\partial u}{\partial t}
&=
-\mathbf{V}\cdot\nabla u
+fv-r_m u+\kappa_u\nabla^2u,\\
\frac{\partial v}{\partial t}
&=
-\mathbf{V}\cdot\nabla v
-fu-r_m v+\kappa_v\nabla^2v .
\end{aligned}
\end{equation}
Here, $f=2\Omega\sin(\phi)$ is the Coriolis parameter, where $\Omega$ is the Earth angular velocity and $\phi$ is the grid latitude.
The surface friction coefficient $r_m$ and wind diffusion coefficients $\kappa_u,\kappa_v$ are learnable positive parameters.

For the remaining state variables, the same transport-diffusion form gives \emph{pressure transport}, \emph{thermodynamic state}, and \emph{humidity evolution}:
\begin{equation}
\begin{aligned}
\frac{\partial p}{\partial t}
&=
-\mathbf{V}\cdot\nabla p
+\kappa_p\nabla^2p+S_p,\\
\frac{\partial \theta}{\partial t}
&=
-\mathbf{V}\cdot\nabla \theta
+\kappa_\theta\nabla^2\theta+S_\theta,\\
\frac{\partial q}{\partial t}
&=
-\mathbf{V}\cdot\nabla q
+\kappa_q\nabla^2q
+S_q-\lambda_c\max(q-q_s,0).
\end{aligned}
\end{equation}
Here, $S_p$, $S_\theta$, and $S_q$ are closure terms generated from the current state, terrain, and time features to represent unresolved pressure, heat, and moisture effects.
In implementation, they are produced by a two-layer convolutional closure network with terrain-conditioned hidden features and channel-wise output scales.
The saturation specific humidity $q_s$ is computed from temperature and pressure, and $\lambda_c\max(q-q_s,0)$ removes supersaturated moisture.
The diffusion coefficients $\kappa_p,\kappa_\theta,\kappa_q$ and condensation coefficient $\lambda_c$ are learnable positive parameters.

The above evolution equations are used to advance the current surface state through repeated PDE substeps.
As shown in Fig.~\ref{fig:system}(b), given the current field $\mathbf{X}_t$, we set $\mathbf{X}^{(0)}=\mathbf{X}_t$ and choose $K$ such that $K\Delta t$ matches one forecast interval.
At the $r$-th substep, the PDE tendencies are converted into a range-scaled Euler update:
\begin{equation}
\begin{aligned}
    \mathbf{X}^{(r+1)}
    &=
    \mathbf{X}^{(r)}
    +
    \boldsymbol{\gamma}\odot
    \mathrm{Scale}
    \big(
    \Delta t\,\mathcal{F}_{\mathrm{PDE}}(\mathbf{X}^{(r)});
    \mathbf{X}^{(r)}
    \big),\\
    &\hspace{3.9cm} r=0,\ldots,K-1 .
\end{aligned}
\end{equation}
Here, $\Delta t=1800\ \mathrm{s}$, and $\mathcal{F}_{\mathrm{PDE}}(\cdot)$ concatenates the tendencies of all state variables.
Following the range-scaling strategy in WeatherGFT~\cite{xu2024weathergft}, $\mathrm{Scale}(\cdot)$ rescales the Euler update according to the spatial range of each state channel.
The factor $\boldsymbol{\gamma}$ is a learnable channel-wise update factor.
After $K$ substeps, the station-oriented PDE branch outputs $\mathbf{X}_{t+1}^{\mathrm{PDE}}=\mathbf{X}^{(K)}$.

\subsection{Data-Driven Diffusion Branch}
\label{sec:ai_branch}
Although the PDE branch provides explicit surface evolution, the simplified surface PDEs cannot fully cover data-driven motion and residual patterns.
As shown in Fig.~\ref{fig:system}(c), \method{} therefore uses a data-driven diffusion branch to provide adaptive correction on the continuous field.
Given the current field $\mathbf{X}_t$ and the step-wise condition field $\mathbf{B}_t$ containing location, time, and lead-time features, the branch encodes a condition-aware hidden field and then evolves it through motion-guided warping.
This branch starts by generating a candidate state:
\begin{equation}
    \widetilde{\mathbf{X}}_{t}
    =
    \mathbf{X}_{t}
    +
    \tau_c
    \mathcal{A}_{\omega}([\mathbf{X}_{t},\mathbf{B}_{t}]),
\end{equation}
where $\mathcal{A}_{\omega}$ is a convolutional candidate network and $\tau_c$ is a learnable scale.
Then,a state gate controls the correction from $\mathbf{X}_t$ to $\widetilde{\mathbf{X}}_t$, producing the encoded hidden field $\mathbf{H}_t$:
\begin{equation}
    \mathbf{H}_{t}
    =
    \mathbf{X}_{t}
    +
    \mathbf{M}_{t}^{s}\odot
    (\widetilde{\mathbf{X}}_{t}-\mathbf{X}_{t}),
\end{equation}
where $\mathbf{M}_{t}^{s}$ is a sigmoid state gate generated from $\mathbf{X}_{t}$, $\widetilde{\mathbf{X}}_{t}$, and $\mathbf{B}_{t}$.

From $\mathbf{H}_t$, the branch estimates a two-channel motion field:
\begin{equation}
    \mathbf{D}_{t}
    =
    \mathcal{M}_{\omega}([\mathbf{H}_{t},\mathbf{B}_{t}]),
\end{equation}
where $\mathcal{M}_{\omega}$ is a motion network and $\mathbf{D}_t$ denotes horizontal displacement.Implementation details of these networks are provided in the appendix.

The hidden field $\mathbf{H}_t$ is then transported by differentiable bilinear warping:
\begin{equation}
    \mathbf{W}_{t}
    =
    \mathrm{Warp}(\mathbf{H}_{t},\mathbf{D}_{t}),
\end{equation}
where $\mathbf{W}_t$ is the warped field.

Finally, the warped change is combined with a residual correction through a gate:
\begin{equation}
    \mathbf{X}_{t+1}^{\mathrm{AI}}
    =
    \mathbf{H}_{t}
    +
    \mathbf{M}_{t}^{d}\odot
    (\mathbf{W}_{t}-\mathbf{H}_{t})
    +
    \mathbf{R}_{t},
\end{equation}
where $\mathbf{M}_{t}^{d}$ is the warp gate generated from $\mathbf{H}_t$, $\widetilde{\mathbf{X}}_t$, and $\mathbf{B}_t$, and $\mathbf{R}_{t}$ is a learned residual correction generated from $\mathbf{H}_{t}$ and $\mathbf{B}_{t}$.
The output of this branch is the data-driven field forecast $\mathbf{X}_{t+1}^{\mathrm{AI}}$.

\subsection{Adaptive Router}
\label{sec:adaptive_fusion}
As shown in Fig.~\ref{fig:system}(d), the adaptive router integrates the PDE and data-driven branch outputs, and then maps the fused field back to station-level forecasts.
Given $\mathbf{X}_{t+1}^{\mathrm{PDE}}$ and $\mathbf{X}_{t+1}^{\mathrm{AI}}$, the router first estimates a fusion weight:
\begin{equation}
    \boldsymbol{\alpha}_{t}
    =
    \sigma\!\left(
    \mathcal{R}_{\rho}
    ([\mathbf{X}_{t+1}^{\mathrm{AI}},
      \mathbf{X}_{t+1}^{\mathrm{PDE}},
      \mathbf{B}_{t}])
    \right),
\end{equation}
where $\mathcal{R}_{\rho}$ is the router network, $\sigma(\cdot)$ is the sigmoid function, and $\boldsymbol{\alpha}_{t}$ controls the contribution of the data-driven branch.
The fused field is then obtained by
\begin{equation}
    \mathbf{X}_{t+1}
    =
    \boldsymbol{\alpha}_{t}\odot\mathbf{X}_{t+1}^{\mathrm{AI}}
    +
    (1-\boldsymbol{\alpha}_{t})\odot\mathbf{X}_{t+1}^{\mathrm{PDE}} .
\end{equation}

The fused field is then sampled at station locations and refined by a unified station residual head:
\begin{equation}
    \hat{\mathbf{S}}_{t+1}
    =
    \Psi^{-1}
    \left(
    \mathrm{Readout}(\mathbf{X}_{t+1})
    +
    \mathcal{H}_{\xi}
    (\Psi(\mathbf{S}_{t-T_{\mathrm{in}}:t-1}),\mathbf{B}_{t})
    \right).
\end{equation}
Here, $\mathrm{Readout}(\cdot)$ denotes sampling the continuous field at station locations, and $\mathcal{H}_{\xi}$ is a unified station residual head shared across variables.
It is implemented as a lightweight MLP with two SiLU hidden layers, which takes the historical station state sequence, station geographic features, time features, and lead step as input and outputs channel-wise scaled station residuals for all state channels.
The inverse transform $\Psi^{-1}(\cdot)$ converts the corrected physical state back to observable weather variables.
Repeating this process autoregressively gives the final multi-station forecast
$\hat{\mathbf{S}}_{t:t+T_{\mathrm{out}}-1}$.

\input{paper/5.1_Experiments_Maintable}

\subsection{Model Training}
\label{sec:model_training}
We train \method{} end-to-end with station-level supervision.
The main loss is computed in the normalized physical state space:
\begin{equation}
    \mathcal{L}_{\mathrm{state}}
    =
    \frac{1}{\sum_{c=1}^{C} w_c}
    \sum_{c=1}^{C}
    w_c\,
    \left\|
    [\Psi(\hat{\mathbf{S}})]_{\mathrm{norm}}^{(c)}
    -
    [\Psi(\mathbf{S})]_{\mathrm{norm}}^{(c)}
    \right\|_1 ,
\end{equation}
where $w_c$ is the state-channel weight.

To align training with observable weather variables, we additionally use an auxiliary loss in the observable variable space:
\begin{equation}
    \mathcal{L}_{\mathrm{obs}}
    =
    \frac{1}{\sum_{j=1}^{C} v_j}
    \sum_{j=1}^{C}
    v_j\,
    \left\|
    \frac{[\hat{\mathbf{S}}]^{(j)}-[\mathbf{S}]^{(j)}}{s_j}
    \right\|_1 ,
\end{equation}
where $v_j$ is the observable-variable weight, and $s_j$ is the batch-wise target standard deviation used to balance variables with different units.
The total objective is
\begin{equation}
    \mathcal{L}
    =
    \mathcal{L}_{\mathrm{state}}
    +
    \lambda_{\mathrm{obs}}\mathcal{L}_{\mathrm{obs}},
\end{equation}
where $\lambda_{\mathrm{obs}}$ controls the strength of the auxiliary observable-space loss.
All neural modules, adaptive router parameters, station residual head, and learnable PDE coefficients are optimized jointly.

%% file: paper/5.1_Experiments_Maintable.tex
\begin{table*}[t]
    \centering
    \renewcommand{\arraystretch}{1.3}
    \footnotesize
    \setlength{\tabcolsep}{2.5pt}
    \resizebox{\textwidth}{!}{
    \begin{tabular}{@{}c|l|c|cccccccc@{}}
        \hline
        \multicolumn{2}{c|}{Models}
        & \makecell{\method{} \\ (Ours)}
        & CDPNet
        & TimeFilter
        & \makecell{MultiPatch\\Former}
        & TimeXer
        & TimeMixer
        & iTransformer
        & Corrformer
        & DLinear \\
        \hline

        \multirow{5}{*}{\rotatebox{90}{weather2k}}
        & pressure
        & \best{4.413}\std{0.090} & 9.840\std{0.644} & 8.625\std{0.137} & 6.750\std{0.122} & 7.738\std{0.080} & 7.684\std{0.119} & 7.700\std{0.077} & \second{5.732}\std{0.047} & 11.317\std{0.129} \\
        \cline{2-11}

        & temp
        & \best{4.607}\std{0.055} & \second{5.856}\std{0.101} & 7.058\std{0.022} & 6.694\std{0.038} & 7.121\std{0.020} & 7.653\std{0.020} & 7.118\std{0.015} & 5.897\std{0.016} & 8.356\std{0.006} \\
        \cline{2-11}

        & humidity
        & \best{127.240}\std{0.807} & 150.312\std{2.075} & 171.486\std{0.824} & 168.889\std{0.440} & 165.972\std{0.204} & 174.633\std{0.756} & 171.770\std{0.373} & \second{148.102}\std{0.213} & 187.482\std{0.078} \\
        \cline{2-11}

        & u-wind
        & \best{2.155}\std{0.007} & \second{2.234}\std{0.011} & 2.597\std{0.002} & 2.573\std{0.004} & 2.516\std{0.005} & 2.581\std{0.006} & 2.606\std{0.003} & 2.374\std{0.003} & 2.747\std{0.001} \\
        \cline{2-11}

        & v-wind
        & \best{2.194}\std{0.008} & \second{2.322}\std{0.013} & 2.820\std{0.010} & 2.789\std{0.009} & 2.768\std{0.005} & 2.864\std{0.013} & 2.863\std{0.006} & 2.534\std{0.003} & 3.083\std{0.001} \\
        \hline

        \multirow{4}{*}{\rotatebox{90}{metonet}}
        & temp
        & \best{4.298}\std{0.095} & \second{4.756}\std{0.077} & 5.348\std{0.069} & 5.310\std{0.030} & 5.389\std{0.067} & 6.298\std{0.037} & 5.314\std{0.039} & 7.444\std{0.164} & 6.762\std{0.046} \\
        \cline{2-11}

        & humidity
        & \best{87.115}\std{0.941} & \second{88.562}\std{1.293} & 96.163\std{0.213} & 92.915\std{0.170} & 94.075\std{1.054} & 102.617\std{0.316} & 95.460\std{0.290} & 113.500\std{0.760} & 103.519\std{0.182} \\
        \cline{2-11}

        & u-wind
        & \best{4.241}\std{0.087} & \second{4.429}\std{0.055} & 4.697\std{0.031} & 4.875\std{0.012} & 4.703\std{0.028} & 5.391\std{0.013} & 4.832\std{0.009} & 5.255\std{0.106} & 5.359\std{0.012} \\
        \cline{2-11}

        & v-wind
        & \best{4.316}\std{0.066} & \second{4.462}\std{0.033} & 5.104\std{0.028} & 5.314\std{0.030} & 5.365\std{0.023} & 5.706\std{0.018} & 5.311\std{0.029} & 5.964\std{0.107} & 5.708\std{0.014} \\
        \hline
    \end{tabular}
    }
    \caption{Experiment results (MSE) on the Weather2K and MeteoNet datasets. All models use 48 hours of historical observations to forecast the next 24 hours. Best results are marked in \best{bold}, and second-best results are \second{underlined}. Complete MAE results are reported in the appendix.}
    \label{tab:weather2k_metonet_results}
\end{table*}

%% file: paper/5_Experiments.tex
\section{Experiments}
\label{sec:eval}
\subsection{Experimental Settings}
\myparagraph{Dataset.}
We evaluate \method{} on two real-world multi-station multivariate weather datasets.
The Weather2K dataset~\cite{Zhu2023weather2K} contains $1{,}866$ ground weather stations across China from January 1, 2017 to August 31, 2021, with 3-hour temporal resolution.
Each station provides observable surface weather variables $[u,v,p,T,RH]$, including air pressure, air temperature, relative humidity, and horizontal wind components converted from wind observations.
The MeteoNet dataset~\cite{meteonet2020} provides hourly surface station observations in northwestern France from January 1, 2016 to December 31, 2018.
After filtering stations with high missing rates, we retain $133$ stations and use the same types of surface weather variables as Weather2K, except that pressure is excluded from the final evaluation because MeteoNet records pressure reduced to sea level and its missing rate reaches $64\%$.
For both datasets, we chronologically split the sequences into training, validation, and test sets with a $6{:}1{:}3$ ratio.

\input{paper/5.1_Experiments_Maintable_different_time}
\input{paper/5.1_Experiments_ablationtable}
\myparagraph{Baselines.}
We compare \method{} with representative baselines from two groups: multi-station weather forecasting models, including Corrformer~\cite{wu2023Corrformer} and CDPNet~\cite{xu2025CDPNet}; and recent general multivariate time series forecasting models, including DLinear~\cite{zeng2023dlinear}, iTransformer~\cite{liu2024itransformer}, TimeMixer~\cite{wang2024timemixer}, TimeXer~\cite{wang2024timexer}, MultiPatchFormer~\cite{naghashi2025multipatchformer}, and TimeFilter~\cite{zhang2025timefilter}.
\myparagraph{Implementation details.}
Except for Corrformer and CDPNet, all baselines are implemented based on Time-Series-Library~\cite{wang2024tssurvey}.
Corrformer and CDPNet are implemented following the hyperparameter settings recommended in their original papers.
For the MeteoNet dataset, pressure is recorded as pressure reduced to sea level and has a missing rate of $64\%$.
Therefore, we exclude pressure from the final evaluation, but estimate a surface pressure reference from station altitude using the standard atmosphere approximation to drive PDE evolution.
For \method{}, the grid height $H$ is set to $32$, and the grid width is automatically adjusted according to the regional aspect ratio.
The PDE substep is fixed as $\Delta t=1800$ seconds, with $K=6$ for Weather2K and $K=2$ for MeteoNet to match their forecast intervals.
The observable-variable auxiliary loss weight is set to $\lambda_{\mathrm{obs}}=0.2$.
We train all trainable models for $10$ epochs using AdamW with a learning rate of $1\times10^{-4}$.
Each experiment is repeated with five random seeds, and we report the mean and standard deviation.
Other hyperparameters, including channel-wise loss weights, are provided in the appendix.
All experiments were conducted on a single NVIDIA RTX 4090 GPU (24GB).

\subsection{Result Analysis}
\myparagraph{Main results.}
Table~\ref{tab:weather2k_metonet_results} reports the main results on Weather2K and MeteoNet.
Overall, \method{} achieves the best performance across all evaluated variables and both datasets in terms of MSE, demonstrating the effectiveness of introducing station-oriented surface PDE evolution into multi-station multivariate forecasting.
Compared with the strongest baseline for each variable, \method{} reduces MSE by about $9.6\%$ on average, showing consistent gains across different datasets and weather variables.
Compared with CDPNet, the closest station-to-field baseline, \method{} brings particularly clear improvements on Weather2K pressure and humidity, reducing MSE from $9.840$ to $4.413$ and from $150.312$ to $127.240$, respectively.
This indicates that terrain-aware field construction and surface PDE evolution are especially beneficial for variables that are sensitive to terrain effects and cross-variable physical coupling.
For wind variables, \method{} also consistently outperforms CDPNet, showing that the proposed PDE branch and adaptive fusion provide gains beyond data-driven continuous field modeling.
Complete MAE results show similar trends and are reported in the appendix.

\myparagraph{Performance across forecast horizons.}
Table~\ref{tab:meteonet_horizon_results} evaluates longer forecast horizons on MeteoNet.
When the forecast horizon extends to $48$ and $96$ hours, \method{} achieves the best MSE on all variables.
This suggests that \method{} is robust under longer forecast horizons, where accumulated errors make station-level forecasting more challenging.

\subsection{Ablation Study}
Table~\ref{tab:ablation} reports the ablation results on Weather2K.
All ablation variants perform worse than the full model, confirming the contribution of each component.
\textbf{w/o PDE} removes the station-oriented surface PDE branch and causes clear degradation, especially on humidity and wind, showing the importance of explicit surface evolution.
\textbf{w/o Data-Driven} leads to the largest overall degradation, indicating that the data-driven branch is necessary for capturing residual dynamics beyond the PDE process.
\textbf{w/o Station Residual} worsens performance on most variables, suggesting that station-level residual correction remains important after sampling from the fused field.
\textbf{w/o $\mathcal{L}_{\mathrm{obs}}$} yields smaller but consistent degradation, showing that observable-variable supervision helps align the learned physical state with weather observations.

\begin{figure}[t]
    \centering
    \includegraphics[width=\linewidth]{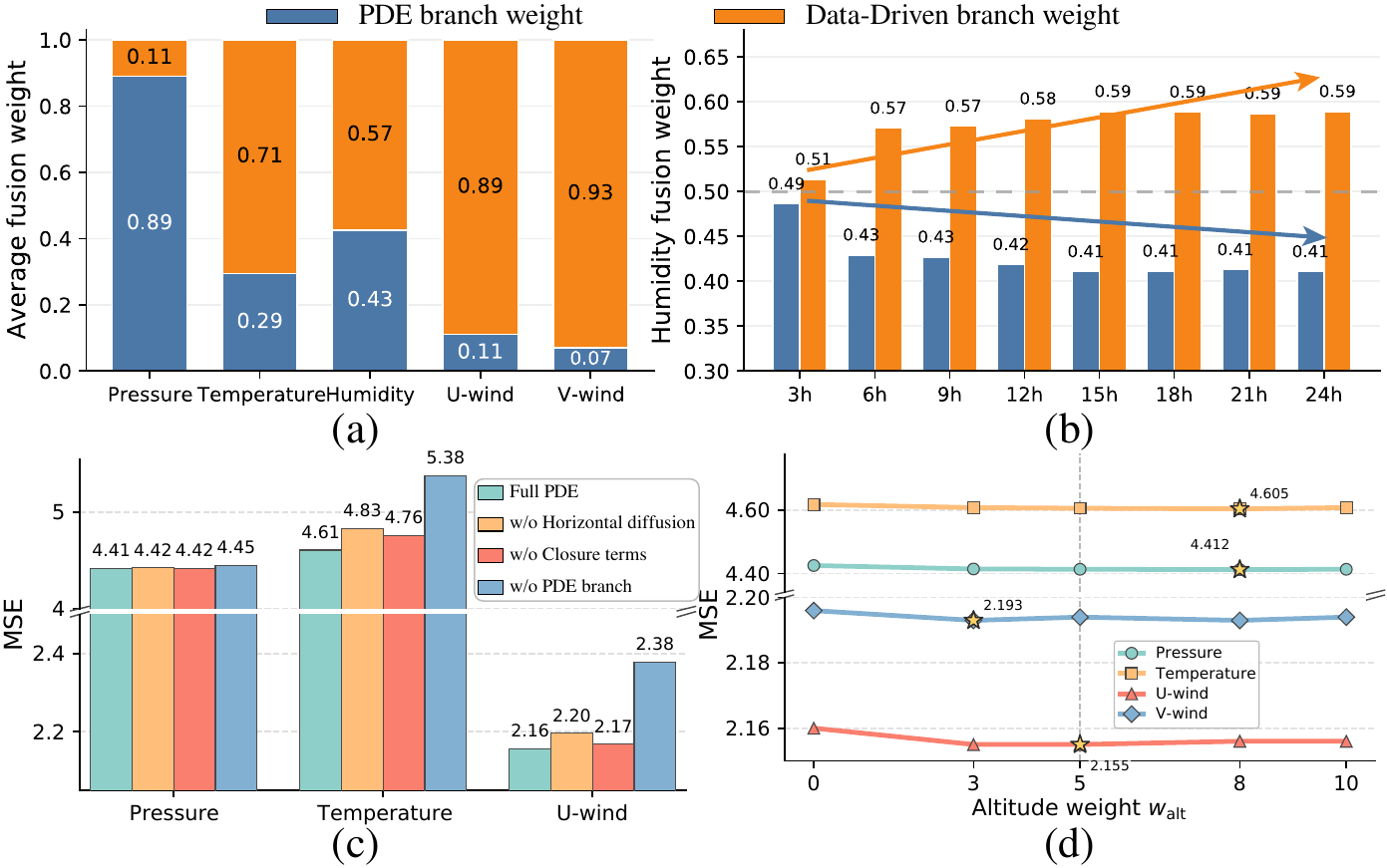}
    \caption{
    (a) Average fusion weights of the PDE branch and data-driven branch for different weather variables.
    (b) Lead-time-wise fusion weights for humidity from 3h to 24h.
    (c) Relative MSE increase after removing key PDE components.
    (d) Sensitivity to the altitude weight $w_{\mathrm{alt}}$ in terrain-aware surface field initialization.
    }
    \label{fig:visualization}
\end{figure}
\subsection{Visualization Study}
Fig.~\ref{fig:visualization} visualizes the adaptive router weights of the two branches.
As shown in Fig.~\ref{fig:visualization}(a), the router assigns variable-specific fusion weights: pressure relies more on the PDE branch, temperature and humidity use both branches, while wind variables rely more on the data-driven branch.
This suggests that the router adaptively balances explicit surface evolution and learned correction according to variable characteristics.
Fig.~\ref{fig:visualization}(b) shows the lead-time-wise weights for humidity, where the data-driven weight gradually increases with the forecast horizon.
This indicates that longer forecasts require stronger adaptive correction to handle accumulated uncertainty.
Fig.~\ref{fig:visualization}(c) further analyzes the PDE components.
Removing horizontal diffusion or closure terms increases errors on most variables, while removing the whole PDE branch causes the largest degradation, especially for temperature and wind.
This supports the contribution of both upper-air inference and residual surface forcing in the surface PDE process.
Fig.~\ref{fig:visualization}(d) shows the sensitivity to the altitude weight $w_{\mathrm{alt}}$.
The performance is stable around $w_{\mathrm{alt}}=5$, which is used in our experiments, indicating that a moderate altitude contribution helps initialize the continuous surface field.

%% file: paper/5.1_Experiments_Maintable_different_time.tex
\begin{table*}[t]
    \centering
    \renewcommand{\arraystretch}{1.3} 
    
    \footnotesize
    \setlength{\tabcolsep}{2.5pt} 
    \resizebox{\textwidth}{!}{
    \begin{tabular}{@{}l|c|c|cccccccc@{}}
        \hline
        \multicolumn{2}{c|}{Models}
        & \makecell{\method{} \\ (Ours)}
        & CDPNet
        & TimeFilter
        & \makecell{MultiPatch\\Former}
        & TimeXer
        & TimeMixer
        & iTransformer
        & Corrformer
        & DLinear \\
        \hline
        
        \multirow{2}{*}{temp}
        & 48 
        & \best{6.978}\std{0.179} & \second{7.288}\std{0.057} & 8.556\std{0.183} & 8.397\std{0.120} & 8.483\std{0.055} & 9.274\std{0.091} & 8.323\std{0.075} & 10.449\std{0.374} & 9.969\std{0.136} \\
        
        & 96 
        & \best{9.849}\std{0.278} & \second{10.685}\std{0.144} & 12.891\std{0.154} & 12.845\std{0.148} & 12.726\std{0.118} & 13.352\std{0.065} & 12.761\std{0.051} & 14.452\std{0.188} & 13.940\std{0.063} \\
        \hline

        \multirow{2}{*}{humidity}
        & 48 
        & \best{103.910}\std{3.200} & \second{107.508}\std{0.805} & 122.763\std{1.142} & 117.479\std{0.315} & 118.568\std{0.778} & 126.363\std{0.822} & 120.362\std{0.260} & 135.347\std{1.055} & 122.470\std{0.283} \\
        
        & 96 
        & \best{133.674}\std{2.090} & \second{138.263}\std{0.455} & 146.440\std{2.454} & 141.550\std{0.817} & 142.138\std{1.698} & 148.828\std{0.826} & 144.545\std{0.504} & 155.502\std{1.214} & 139.269\std{0.108} \\
        \hline

        \multirow{2}{*}{u-wind}
        & 48 
        & \best{5.794}\std{0.081} & \second{5.865}\std{0.039} & 6.669\std{0.118} & 6.686\std{0.020} & 6.630\std{0.014} & 7.137\std{0.018} & 6.724\std{0.025} & 7.138\std{0.076} & 6.669\std{0.025} \\
        
        & 96 
        & \best{7.174}\std{0.225} & \second{7.219}\std{0.037} & 8.433\std{0.172} & 8.453\std{0.033} & 8.513\std{0.032} & 8.970\std{0.024} & 8.480\std{0.028} & 9.028\std{0.307} & 7.881\std{0.007} \\
        \hline

        \multirow{2}{*}{v-wind}
        & 48 
        & \best{5.966}\std{0.053} & \second{5.991}\std{0.012} & 7.110\std{0.135} & 7.100\std{0.031} & 7.235\std{0.009} & 7.388\std{0.042} & 7.109\std{0.056} & 7.693\std{0.083} & 6.980\std{0.027} \\
        
        & 96 
        & \best{7.178}\std{0.150} & \second{7.227}\std{0.014} & 8.649\std{0.161} & 8.788\std{0.038} & 9.001\std{0.036} & 9.063\std{0.031} & 8.674\std{0.042} & 9.210\std{0.195} & 8.087\std{0.007} \\
        \hline
    \end{tabular}
    }
    \caption{Experiment results (MSE) on the MeteoNet dataset with different forecast horizons (48h and 96h). All models use 48 hours of historical observations. Best results are marked in \best{bold}, and second-best results are \second{underlined}. Complete MAE results are reported in the appendix.}
    \label{tab:meteonet_horizon_results}
\end{table*}

%% file: paper/5.1_Experiments_ablationtable.tex
\begin{table}[t]
    \centering
    \begingroup
    \footnotesize
    \setlength{\tabcolsep}{0.5pt}
    \renewcommand{\arraystretch}{1.05}
    \resizebox{\linewidth}{!}{
    \begin{tabular}{@{}lccccc@{}}
        \toprule
        Method & pressure & temp & humidity & u-wind & v-wind \\
        \midrule
        Full / \method{}
        & \best{4.41}\std{0.09}
        & \best{4.61}\std{0.06}
        & \best{127.24}\std{0.81}
        & \best{2.16}\std{0.01}
        & \best{2.19}\std{0.01} \\
        w/o PDE
        & 4.56\std{0.03}
        & 5.42\std{0.06}
        & 150.62\std{1.55}
        & 2.39\std{0.01}
        & 2.41\std{0.00} \\
        w/o Data-Driven
        & 4.84\std{0.06}
        & 5.87\std{0.11}
        & 161.17\std{0.70}
        & 2.41\std{0.06}
        & 2.42\std{0.01} \\
        w/o Station Residual
        & 4.72\std{0.06}
        & 5.36\std{0.11}
        & 148.95\std{0.74}
        & 2.38\std{0.01}
        & 2.40\std{0.00} \\
        w/o $\mathcal{L}_{\mathrm{obs}}$
        & 4.45\std{0.07}
        & 4.76\std{0.01}
        & 134.88\std{0.61}
        & 2.16\std{0.01}
        & 2.20\std{0.01} \\
        \bottomrule
    \end{tabular}
    }
    \endgroup
    \caption{Ablation study results (MSE) on the Weather2K dataset. All models use 48 hours of historical observations to forecast the next 24 hours. Best results are marked in \best{bold} based on higher-precision values.}
    \label{tab:ablation}
\end{table}

%% file: paper/6_Conclusion.tex
\section{Conclusion}
\label{sec:conclusion}
We proposed \method{}, a station-oriented surface PDE learning model for multi-station multivariate weather forecasting.
\method{} lifts station observations into a terrain-aware continuous field, evolves it with surface PDE processes, and integrates data-driven correction through an adaptive router.
Experiments on real-world datasets demonstrate its effectiveness over representative baselines.